\documentclass[letterpaper]{article} 
\usepackage{aaai23}  
\usepackage{times}  
\usepackage{helvet}  
\usepackage{courier}  
\usepackage[hyphens]{url}  
\usepackage{graphicx} 
\usepackage{natbib}  
\usepackage{caption} 
\usepackage{amssymb}
\usepackage{amsfonts}
\usepackage{xcolor}
\usepackage[hidelinks]{hyperref}
\usepackage{booktabs}
\usepackage{amsmath}
\usepackage{algorithm}
\usepackage{algorithmic}
\usepackage{bm}
\usepackage[figuresright]{rotating}
\usepackage{multirow}
\usepackage{adjustbox}

\newcommand{\eg}{\textit{e.g.}}

\newcommand{\ie}{\emph{i.e.}}
\newcommand{\mc}{\mathcal}
\newcommand{\tb}{\textbf}
\newcommand{\td}{\tilde}
\newcommand{\ti}{\textit}
\newcommand{\ul}{\underline}
\newcommand{\bs}{\boldsymbol}

\newcommand{\mb}{\mathbb}

\newcommand{\tabincell}[2]{\begin{tabular}{@{}#1@{}}#2\end{tabular}}

\usepackage{newfloat}
\usepackage{listings}
\DeclareCaptionStyle{ruled}{labelfont=normalfont,labelsep=colon,strut=off} 
\floatstyle{ruled}
\newfloat{listing}{tb}{lst}{}
\floatname{listing}{Listing}
\title{Meta-Auxiliary Learning for Adaptive Human Pose Prediction}
\author{
    Qiongjie Cui\textsuperscript{\rm 1},\,\,
    Huaijiang Sun\textsuperscript{\rm 1}\footnote{indicates corresponding author},\,\,
    Jianfeng Lu\textsuperscript{\rm 1},\,\,
    Bin Li\textsuperscript{\rm 2},\,\,
    Weiqing Li\textsuperscript{\rm 1}   
}
\affiliations{
    \textsuperscript{\rm 1}Nanjing University of Science and Technology\\
    \textsuperscript{\rm 2}Tianjin AiForward Science and Technology Co., Ltd., China\\

    cuiqiongjie@njust.edu.cn, sunhuaijiang@njust.edu.cn
    \\
    lujf@njust.edu.cn,
    libin@aiforward.com,
    li\_weiqing@njust.edu.cn
}

\usepackage{bibentry}

\begin{document}

\maketitle
\begin{abstract}
  Predicting high-fidelity future human poses, from a historically observed sequence, is decisive for intelligent robots to interact with humans.
  Deep end-to-end learning approaches, which typically train a generic pre-trained model on external datasets and then directly apply it to all test samples, emerge as the dominant solution to solve this issue.
  Despite encouraging progress, they remain non-optimal, as the unique properties (\eg, motion style, rhythm) of a specific sequence cannot be adapted.
  More generally, at test-time, once encountering unseen motion categories (out-of-distribution), the predicted poses tend to be unreliable.
  Motivated by this observation, we propose a novel test-time adaptation framework that leverages two self-supervised auxiliary tasks to help the primary forecasting network adapt to the test sequence.
  In the testing phase, our model can adjust the model parameters by several gradient updates to improve the generation quality.
   However, due to catastrophic forgetting, both auxiliary tasks typically tend to the low ability to automatically present the desired positive incentives for the final prediction performance.
  For this reason, we also propose a meta-auxiliary learning scheme for better adaptation.
  In terms of general setup, our approach obtains higher accuracy, and under two new experimental designs for  out-of-distribution data (unseen subjects and categories), achieves significant improvements.
  \vspace{-.5em}
\end{abstract}

\newcommand{\predy}{\td{\tb{Y}}_{1:\Delta T}}
\newcommand{\gty}{\tb{Y}_{1:\Delta T}}
\newcommand{\predx}{\td{\tb{X}}_{1:T}}
\newcommand{\missingx}{\hat{\tb{X}}_{1:T}}
\newcommand{\gtx}{\tb{X}_{1:T}}
\newcommand{\apy}{\td{\ti{\tb{Y}}}}

\section{Introduction}
Human pose forecasting, accurately predicting how a person will move in the near future, is a fundamental task in computer vision, which has enormous potential in machine intelligence, and human-robot interaction \cite{Gui2018AdversarialGH,wang2021pvred,liu2021aggregated,piergiovanni2020adversarial,martinez2021pose,sofianos2021space}.

\begin{figure}[t]
  \centering
  \includegraphics[width=3.2in,height=1.65in]{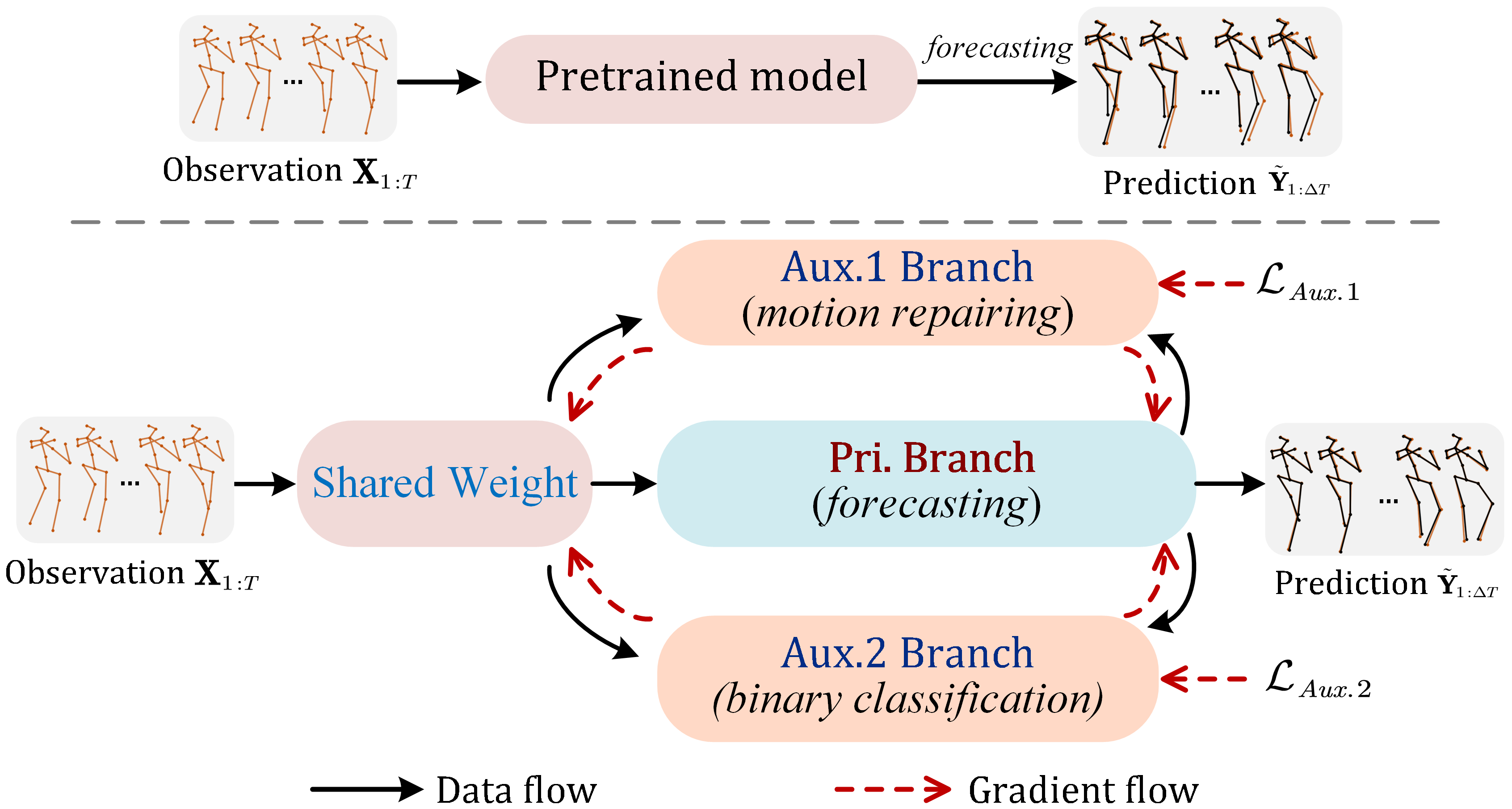}
  \caption{
    Comparison of a classic deep end-to-end model (top) with our approach (bottom) at test-time.
    Given an observed sequence $\gtx$, typical approaches indiscriminately utilize the pre-trained model obtained from large-scale datasets to generate the prediction $\predy$, which is sub-optimal, as the internal information within the specific sample is ignored.
    By contrast, in the test phase, our model learns to adapt to the unique properties of the test sample via several meta-learning steps.
    Here, the black poses denote the prediction, and the underlying orange ones are the GT.
  }
  \vspace{-.5em}
  \label{f1}
\end{figure}

Over the past few years, extensive literature has sprung up exploring this fascinating topic, with deep-learning based end-to-end approaches proving increasingly popular \cite{li2020multitask,Gui2018FewShotHM,li2018convolutional,li2021symbiotic}.
Researchers are prone to train on external large-scale datasets \cite{Ionescu2014Human36MLS} to achieve a generic pre-trained model, which is then indiscriminately applied to all test sequences with the same set of network weights in the inference stage \cite{Jain2016StructuralRNNDL,mao2021generating,dang2021msr}.
These approaches have extensively investigated this issue from various perspectives, emerging as the mainstream solutions.

While the empirical result is encouraging, it is not optimal.
In real-world applications, the inherent defects of the existing DNN-based models \cite{Ruiz2018HumanMP,Gopalakrishnan2018ANT,barsoum2018hp,kundu2018bihmp,aliakbarian2021contextually} cannot be overlooked, among which the major one is that the features learned from external datasets can hardly cover the unique attributes within a specific sequence, including motion style and rhythm, as well as the height and physical proportion of actors, etc.
More generally, for diverse human motion, large-scale datasets have a low ability to cover all categories, which means that, at test time, unseen action categories are frequently encountered \cite{ulyanov2018deep,shin2021test,hao2021test}.
In this case, the model tends to focus on the dominant distribution in training, while failing to take account of the unique patterns of new action categories (out-of-distribution), therefore, unreliable results may be yielded.
This inability to adapt to the internal properties of a given sequence hinders the realistic application of the predictive algorithms \cite{yuan2020dlow,mao2020history,shu2021spatiotemporal}.

To solve it, we propose a novel Test-Time Adaptation (TTA) approach.
Concretely, our model falls into auxiliary learning, where the network consists of one primary task and two self-supervised auxiliary ones.
The primary task (Pri.) focuses on mapping historical observations to the predicted poses.
The auxiliary task-1 (Aux.1) is a simple binary classifier to distinguish whether the input sequence is a scrambled counterpart of the observation.
As a contrast, some joints of the observed sequence are randomly removed to construct the corrupted sequence, and then Aux.2 aims to repair these missing joints.
The Pri., Aux.1, and Aux.2 share most of the parameters, and are jointly trained to achieve a base model.
Then, in the testing phase, Aux.1 and Aux.2 behave as a regularization to further update the shared weights to enhance the generalization for specific sequences.

Intuitively, the auxiliary task provides rich semantic cues to fine-tune the model parameters \cite{chi2021test,varsavsky2020test}.
However, empirical results show that, a rough update of the base model may lead to the criticized \ti{negative transfer}, as the invalid message may be exchanged \cite{xiao2018gated,vafaeikia2020brief}.
To solve it, we design a Gate Sharing Unit (GSU), which learns to control the relative intensity of message transmission among tasks in both training and testing, to pass the favorable information, while hinder the redundant or even incorrect ones.

Even so, there is a legacy problem: how to ensure that the Pri. branch obtains better adapted parameters to ensure the forecasting performance of specific sequences.
For this purpose, inspired by MXML \cite{liu2019self,chi2021test}, we integrate meta-learning into auxiliary learning to form meta-auxiliary learning.
Our meta-objective is to optimize the whole network via meta-auxiliary learning so that the Pri. branch can better adapt to test sequences.
Note that we call the pair composed of the observed and the future poses the 'task' in the meta-learning nomenclature.
Moreover, for each observed sequence, the adapted parameters are different, and its specific motion patterns can be generalized.

Methodologically, to capture the spatio-temporal pattern of skeleton data, we introduce two virtual relay nodes into the sparse transformer, to form the Spatial Sparse-Relay Transformer (SS-RT) and Temporal Sparse-Relay Transformer (TS-RT) \cite{Child2019GeneratingLS,aksan2020attention,cai2020learning}.
The relay nodes are capable of receiving information from all human joints along with spatial and temporal aspects, to extract the global spatio-temporal correlations.
With the sparse transformer and relay-nodes update, the newly designed SS-RT and TS-RT explicitly consider the human topology and temporal smoothness of motion sequences, as well as long-term correlations in space and time.

Our contributions are multifaceted:
(1) We develop a test-time adaptation approach that leverages meta-auxiliary learning to enable fast and effective adaptation to the specific information within test sequences.
(2) Both motion repairing and binary classification are introduced as our self-auxiliary tasks, which are exploited to automatically optimize the pre-trained model, without any extra manual labeling.
(3) To avoid the negative transfer across multi-tasks, the GSUs are designed to allow valid information to be passed easily among tasks, while preventing useless one.
(4) 
On two widely-used benchmarks, our model achieves state-of-the-art performance, and under out-of-distribution data, outperforms the existing methods by a large margin.
To our knowledge, this is the first attempt to improve the prediction quality for unseen categories and subjects in the real world.

\section{Related Work}
\textbf{Human Motion Forecasting.}
Deep end-to-end learning approaches have dominated this issue, with the attraction of providing high flexibility and exceptional performance \cite{martinez2017human,corona2020context,cui2020learning,li2020multitask}.
Researchers typically regard human motion forecasting as a special seq2seq generation problem, and propose a variety of RNN variants to extract the temporal pattern of 3D skeleton sequences \cite{Tang2018LongTermHM,Gui2018FewShotHM,Chiu2018ActionAgnosticHP}, which have yielded promising results.
Despite this, due to the error accumulation and the failure of accessing the topological relationship, the predicted frame tends to converge to an unexpected and static pose \cite{fragkiadaki2015recurrent,Jain2016StructuralRNNDL,Gopalakrishnan2018ANT,martinez2017human}.

Nowadays, various GNN-based models are being developed to extract the semantic connectivity of the 3D skeleton sequence, with promising results \cite{mao2019learning,cui2020learning,li2020dynamic,li2020multitask,dang2021msr,ma2022progy,zhong2022spatio}.
However, GCNs are capable only of gathering information from the local neighbor joints, and have a limited capacity to capture long-term relationships.

Currently, researchers attempt to exploit the Transformer to achieve the long-range correlation, whereas, it fails to consider the meaningful topology and temporal smoothness of motion sequences, and brings more computational cost \cite{mao2020history,aksan2020attention,guo2022multi}.
In contrast, our approach, which includes a sparse transformer and virtual relay nodes, allows us to explicitly focus on the meaningful local structure and temporal continuity while still extracting long-term correlations.

In real applications, the above approaches remain a significant limitation, \ie, the specific properties of test sequences cannot be adapted.
This work aims to solve it.

\begin{figure*}[t]
  \centering
  \includegraphics[width=6.9in,height=2.05in]{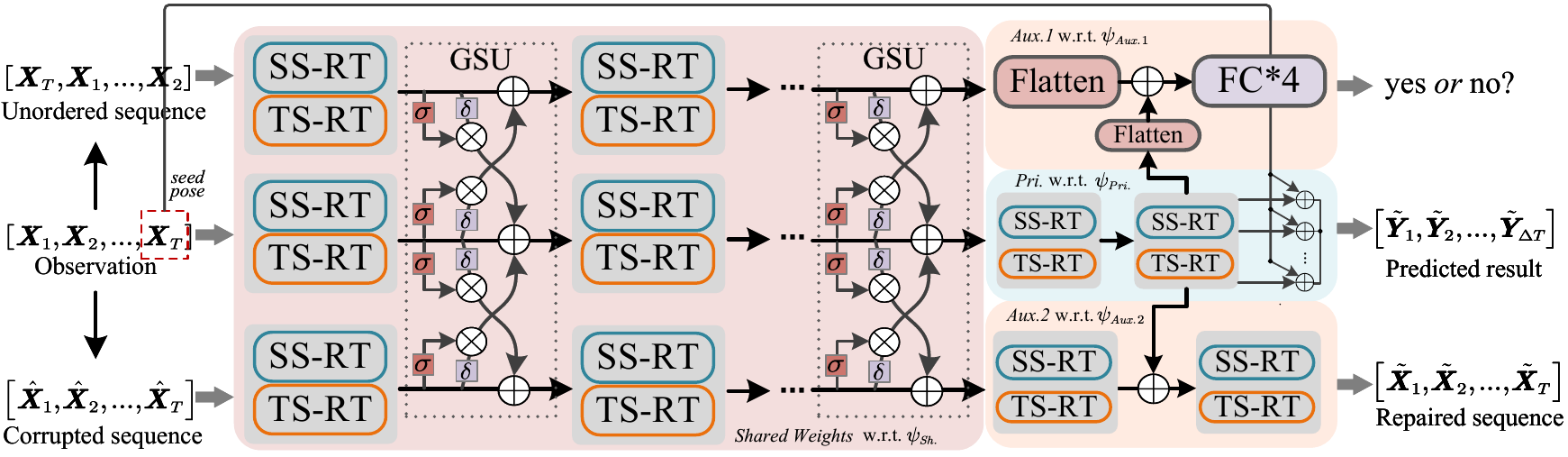}
  \vspace{-.5em}
  \caption{\tb{Illustration of our approach.}
  It involves a primary (Pri.) task and two self-supervised auxiliary (Aux.) ones, sharing most of the parameters w.r.t $\psi_{Sh.}$, except for the task-specific components w.r.t $\{\psi_{Pri.}, \psi_{Aux.1}, \psi_{Aux.2}\}$.
  The Pri. task is concerned with mapping historical observations to the expected prediction.
  The objective of Aux.1 is to provide the correct label of the scrambled sequence, and Aux.2 is to repair the missing joints in the corrupted sequence, where both the scrambled and corrupted sequence are derived from the observation.
  With the proposed GSU, the valid contexts can be exchanged, while the invalid or incorrect ones are blocked.
  $\sigma$, $\delta$ denotes the sigmoid and LeakyReLU function, respectively.
  $\otimes $ is element-wise product and $\oplus$ is addition.
  The last observed frame is regarded as the seed pose (red rectangle).
   }
  \vspace{-.5em}
  \label{f2}
\end{figure*}

\textbf{Test-time Adaptation.}
To improve the generalization for diverse distributions , the test-time adaptation (TTA) scheme is recently proposed \cite{chi2021test,varsavsky2020test,hu2021fully,shin2021test}.
Typically, deep learning algorithms are trained on external datasets to produce a general model, and before making decisions, TTA resorts to auxiliary tasks to neatly fine-tune the weights according to the internal knowledge of test samples.
Due to the utilization of both external and internal information, superior outcomes are achieved \cite{he2021autoencoder,hao2021test}.

However, the existing test-time adaptation technologies remain a key challenge, that is, the auxiliary task may send inaccurate or even incorrect messages to the primary task \cite{vafaeikia2020brief,Cui_2021_CVPR,xiao2018gated}.
To address it, we elaborate a simple but effective gated sharing unit (GSU) that adaptively release the important context while preventing others.

\textbf{Meta-learning.} 
Our work is related to mete-learning (learning to learn), particularly the model-agnostic version (MAML), which allows the pre-trained model to be adjusted to perform the fast adaptation of individual samples. 
\cite{liu2022towards} uses MAML for multi-domain single image dehazing, with the meta-objective of learning consistency across the losses of different tasks.
Along with MAML, \cite{liu2019self} presents the meta-auxiliary learning (MXML) framework, which generates labels for  additional auxiliary tasks.
Inspired by the MXML, \cite{chi2021test} also achieves a fast adaptation to improve the performance of the primary deblurring operation for unseen images.
Our approach, which draws inspiration from these publications in part, involves the following two changes: we design two auxiliary tasks to identify more effective semantics; our auxiliary tasks are self-supervised for the automatic inference.

\section{Proposed Approach}
Suppose that $\gtx = [\bs{X}_{1},\bs{X}_{2},...,\bs{X}_{T}]$ is an observed sequence over horizon $T$, where each $\bs{X}_t = [\bs{j}_1, \bs{j}_2, ..., \bs{j}_N]\in \mb{R}^{N\times D}$ records the 3D coordinate of $N$ human joints in a frame.
Current DNN-based models prone to directly train a mapping from the observation to the future sequence, $\mc{M}:\gtx\rightarrow \gty$, with $\gty =\{\bs{Y}_{1},\bs{Y}_{2},...,\bs{Y}_{\Delta T} \} $.

In contrast, our approach incorporates the following developments.
(1) Two self-auxiliary tasks are introduced, sharing the majority of model weights and allowing collaborative training alongside the primary forecasting one.
Additionally, both Aux. tasks are connected to the Pri. branch, and therefore the effective semantic clues can be provided as a high-order regularization.
(2) To avoid negative transfer across tasks, we build the GSU to prevent the passage of erroneous/incorrect messages.
(3) We first train on large-scale datasets to achieve a base model, and the ultimate goal is to further optimize it at test-time, to automatically adapt to the sample-specific properties, and then yield more realistic predicted results $\predy = \{\apy_{1},\apy_{2},...,\apy_{\Delta T} \}$.
(4) In practical studies, the naive updates might not bring desired improvements. 
To solve it, a meta-auxiliary learning framework is proposed, which learns the better-adapted parameters for the effective test-time adaptation of a specific sequence.

\subsection{Network Architecture}
The network architecture consists of one primary branch and two self-supervised auxiliary ones, as seen in Fig.\ref{f2}.
For convenience, the following uses subscripts to indicate the spatial indexes, and superscripts for temporal indexes.

\tb{Primary Branch.} The Pri. is intended to predict future motions, where its backbone comprises SS-RT and ST-RT to extract the spatio-temporal correlation of motion sequences.

\ti{Spatial Sparse-Relay Transformer (SS-RT)} is implemented to capture the spatial correlation.
In contrast to the vanilla version \cite{vaswani2017attention}, we use the spatial sparse transformer (SST) to explicitly consider the skeletal structure \cite{Child2019GeneratingLS}.
Moreover, we attach a virtual spatial-relay vertex, which utilizes a separate transformer, called spatial-relay transformer (SRT), to directly aggregate the global information in a frame, and distribute it to each one to consider the long-term correlation.

Let $\tb{c}^t = \{ \bs{c}_1^t,\bs{c}_2^t,...,\bs{c}_N^t \} \in \mb{R}^{N\times C_{in}}$ be the feature at $t$-th frame, and $\bs{c}^{t}_{r}$ be a spatial-relay vertex.
For each node $\bs{c}_i^t$, we use 3 linear transformations to generate a query $\bs{q}_i\in \mb{R}^{d}$, a key $\bs{k}_i\in \mb{R}^{d}$ and a value $\bs{v}_i\in \mb{R}^{d}$.
The SST is used to consider the natural connectivity of the human skeleton:
\begin{equation}
  \bs{c}_i^{\prime t}= \sum softmax (\frac{\bs{q}_i \cdot \bs{k}_j}{\sqrt{d}})\bs{v}_j, j\in \{i,\mc{N}_i,r\},
  \label{e1}
\end{equation}
where $\mc{N}_i$ is the neighbors of $i$th joint, and $r$ stands for the label of the spatial-relay vertex.
Then, the meaningful inductive bias of the human skeleton is expressly considered.

In addition, we make use of SRT to capture the long-term spatial correlation:
\begin{small}
  \begin{equation}
    \bs{c}_r^{\prime t} \!= \sum softmax  (\frac{\bs{q}_r \cdot \bs{k}_j}{\sqrt{d}})\bs{v}_j, j\!\in\! \{r\}\!\cup \!\{j\!:\!1\le j\le N\}.
  \end{equation}  
  \vspace{.6em}
\end{small}

\vspace{-1em}
\noindent By stacking the SST and SRT, our SS-RT is formed, which is capable of extracting the intrinsic connections of human joints, and meanwhile, capturing the long-term spatial correlation at intra-frame.
The resulting output of SS-RT can be formalized as: $\tb{c}^{\prime t} = \{ \bs{c}_1^{\prime t},\bs{c}_2^{\prime t},...,\bs{c}_N^{\prime t} \} \in \mb{R}^{N\times C_{out}}$.

\ti{Temporal Sparse-Relay Transformer (TS-RT)} consists of a temporal sparse transformer (TST) for extracting the local inter-frame smoothness, and a temporal-relay transformer (TRT) for long-term temporal dependency.
Let $\tb{c}_v = \{\bs{c}_v^1,\bs{c}_v^2,...,\bs{c}_v^T\}\in\mb{R}^{T\times C_{in}}$ be the input hidden state, for $v \in N$, with $\bs{c}_v^{i} \in \mb{R}^{C_{in}}$, and $\bs{c}_v^r$ be the feature of temporal-relay node, 3 linear transformations are exploited to produce $\bs{q}_i\in \mb{R}^{d}$, $\bs{k}_i\in \mb{R}^{d}$ and $\bs{v}_i\in \mb{R}^{d}$.
The TST is defined as:
\begin{small}
  \begin{equation}
    \bs{c}_{v}^{\prime i} =\sum softmax (\frac{\bs{q}^{i} \cdot \bs{k}^{j}}{d}) \bs{v}^{j}, j \in\{i, i-1, i+1, r\}.
\end{equation}
\vspace{.6em}
\end{small}
Then, the temporal-relay node is updated with the TRT:
  \begin{small}
\vspace{-.6em}
\begin{equation}
  \bs{c}_v^{\prime r} \! = \! \sum softmax (\frac{\bs{q}^r\cdot \bs{k}^j}{d})\bs{v}^j, j  \!\in   \! \{r\}\cup \{j \!: \!1\le j\le T\}.
  \vspace{.7em}
  \label{e4}
\end{equation}
\end{small}

\vspace{-1em}
\noindent The TST and TRT are stacked to create the TS-RT, where the output feature is $\tb{c}_v^{\prime} = \{ \bs{c}_v^{\prime 1},\bs{c}_v^{\prime 2},...,\bs{c}_v^{\prime T} \} \in \mb{R}^{T\times C_{out}}$.

With the TST and TRT, the TS-RT enables the consideration of both local and global temporal correlation, which is crucial for human motion prediction.
In both SS-RT and TS-RT, we set $d= 64$, and in keeping with recent progress \cite{vaswani2017attention,Devlin2019BERTPO}, we exploit $H=8$ independent heads to stabilize the training.

Finally, as illustrated in Fig.\ref{f2}, the Pri. branch is composed of 9 shared blocks and a task-specific one, each of which is formed by a SS-RT and a TS-RT.
\ti{The detailed illustrations of SS-RT and TS-RT refer to the supplementary material.}

Following the recent works \cite{mao2020history,cui2020learning}, the combination of $L_2$ distance and bone length loss is exploited as the loss of the Pri.:
\begin{small}
  \begin{equation}
    \mc{L}_{Pri.}=\|\gty-\predy\|_{2} +\eta\mc{L}_{B}(\gty, \predy), 
    \end{equation}
  \vspace{.6em}
\end{small}

\vspace{-1em}
\noindent where $\mc{L}_B$ is the function to calculate the bone length difference of two motion sequences, and $\eta=0.04$, as in \cite{Xia2018NonlinearLM,cui2020learning}.
$\predy$ is the predicted poses, $\gty$ is the corresponding GT.

\subsubsection{Auxiliary Branches.}
Our approach involves two self-supervised auxiliary branches: Aux.1 (binary classifier) and Aux.2 (motion repairing), as discussed below.

The Aux.1 is a typical binary classifier whose purpose is to identify whether the input counterpart of the observation sequence is disordered.
In \cite{sun2020test,geirhos2018generalisation}, the auxiliary branch is specified as a simple classification task, which improves the performance of the main task at test-time.
In image processing \cite{varsavsky2020test,shin2021test,he2021autoencoder}, the image plane is rotated at a certain angle and the auxiliary task seeks to anticipate the rotation angle.
Such a classification task has been proven to provide an effective context for primary image analysis.
Besides, \cite{li2020multitask} introduces a multi-task model for human motion prediction, where the auxiliary task is employed to yield the correct category of the observed sequence.
These solutions have demonstrated that the basic classifier is an effective auxiliary task to enhance the primary branch, which motivates us to design our auxiliary task as a simple self-supervised binary classifier.
Let $p$ be the correct label of the disordered sequence, and $\td{p}$ the predicted one. The loss function of Aux.1 can be denoted as:
\begin{equation}
    \mc{L}_{Aux.1} = -\left (p\log \td{p} + \left (1-p \right )\log(1-\td{p})\right ).
    \label{e6}
  \end{equation}

  On the other hand, \cite{Cui_2021_CVPR} proposes to take repairing the missing joints as the auxiliary task of the predictor, which has generated the more realistic prediction result.
  Therefore, we design our Aux.2 as a motion repairing task.
  Specifically, we randomly remove some joints in the observed sequence, and the Aux.2 is exploited to impute these missing values.
  The loss function of Aux.2 is described as:
  \begin{small}
    \begin{equation}
     \mc{L}_{Aux.2}=\|\gtx - \predx\|_{2} +\mu \mc{L}_{B}(\gtx, \predx).
      \end{equation}
      \label{e7}
  \end{small}
  
  \vspace{-1em}
  \noindent We set $\mu = 0.04$.
  $\predx$ is the repaired sequence, $\gtx$ is the underlying complete observation.
  
  Compared with the Pri., the goals of our Aux. branches are rather straightforward and can be swiftly carried out during testing.
  Similar to the Pri., both the Aux.1 and Aux.2 include 9 shared blocks formed by SS-RT and TS-RT, as well as a task-specific portion, as shown in Fig.\ref{f2}.
  
  \begin{figure}[t]
    \centering
    \includegraphics[width=3.in,height=1.2in]{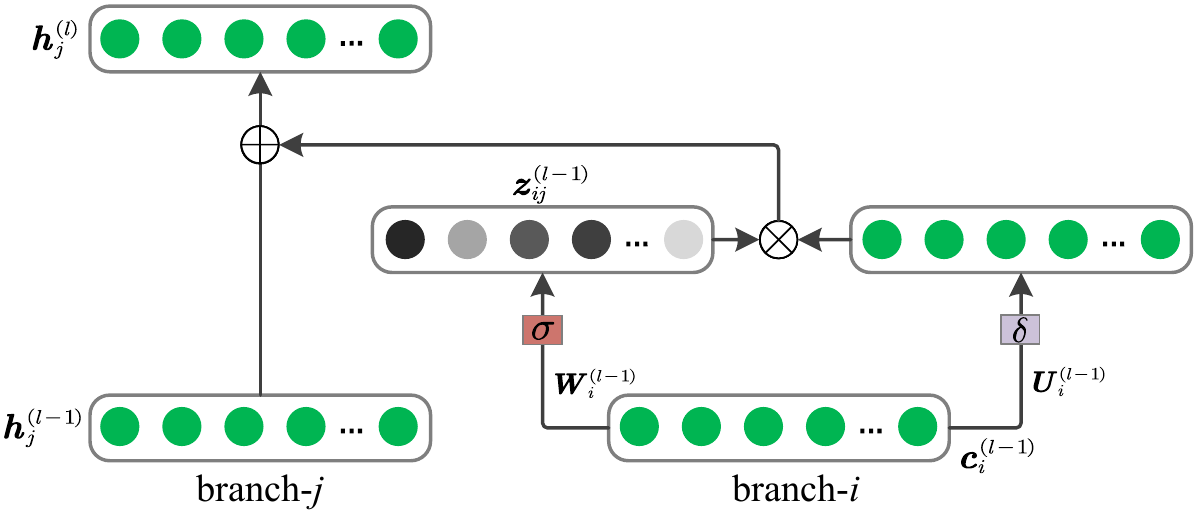}
    \vspace{-.5em}
    \caption{{Illustration of the Gated Sharing Unit.}}
    \vspace{-1em}
    \label{f3}
  \end{figure}

  \subsection{Gated Sharing Units (GSUs)}
  Our network design fits into a multi-task framework in part, where redundant or even incorrect information may be exchanged among tasks, resulting in the criticized negative transfer.
  To mitigate it, we provide a simple but effective GSU to adaptively control the passage of features.
  
  Given a feature $\bs{c}_i^{l-1}\in \mb{R}^{C_{out}}$ of branch-$i$ at $l$-th layer, we first use a single FC layer to achieve the weight:
  \begin{small}
    \vspace{-.3em}
  \begin{equation}
    \bs{z}^{l-1}_{ij} = \sigma(\bs{W}_i^{l-1}\bs{c}_i^{l-1} + \bs{b}_i^{l-1} ),
    \vspace{.2em}
    \label{e8}
  \end{equation}
  \end{small}
  
  \vspace{-1em}
  \noindent where $\sigma$ denotes the sigmoid function, and $\bs{z}^{l-1}_{ij}\in \mb{R}^{C_{out}} $ is the intensity of the message passing form branch-$i$ to branch-$j$.
  $\bs{W}_i^{l-1} \in \mb{R}^{C_{out}\times C_{out}}$ and $\bs{b}_i^{l-1} \in \mb{R}^{C_{out}}$ are the learnable weight and bias, respectively.
  Let $\bs{h}_{j}^{l-1}\in \mb{R}^{C_{out}}$ be the feature at the previous layer of branch-$j$, the output $\bs{h}_j^{l}\in \mb{R}^{C_{out}}$ is then calculated:
  \vspace{-.3em}
  \begin{small}
    \begin{equation}
    \bs{h}^{l}_j = \bs{z}_{ij}^{l-1} \otimes  \delta(\bs{U}_i^{l-1} \bs{c}_i^{l-1} + \bs{e}_{i}^{l-1})+\bs{h}_j^{l-1},
    \vspace{.2em}
    \label{e9}
  \end{equation}
  \end{small}

  \vspace{-1.em}
  \noindent where $\bs{U}_i^{l-1} \in \mb{R}^{C_{out}\times C_{out}}$ is the learnable weight, and $\bs{e}_i^{l-1} \in \mb{R}^{C_{out}}$ is the bias.
  $\otimes$ denotes the element-wise product, and $\delta$ is LeakyReLU with a slope of $0.2$.
Fig.\ref{f3} presents an illustration of the proposed GSU.

\subsection{Joint Training and Meta-auxiliary Learning}
\tb{Joint Training.}
In our model, we have introduced a primary branch and two auxiliary ones.
Since our model involves multiple branches, it can be directly trained, much as the multi-task learning solutions \cite{Cui_2021_CVPR,li2020multitask,chi2021test}.
The overall objective is:
\vspace{-.3em}
\begin{equation}
  \mc{L} = \mc{L}_{Pri.} + \mc{L}_{Aux.1} +\mc{L}_{Aux.2}.
  \label{e10}
  \vspace{-.2em}
\end{equation}
Once the training is complete on the external dataset, the pre-trained model, w.r.t. $\mc{M}_{\psi}$, is attained, which is regarded as the initialization of the meta-auxiliary learning.

\tb{Meta-auxiliary Learning.} Due to the failure of exploiting the internal properties of test samples, the pre-trained model learned from Eq.\ref{e10} has a low ability to adapt to the unseen data.
We solve this problem by using the proposed meta-auxiliary learning to obtain the optimal parameter that is conducive to adapting to a given motion sample.

For each to use, we decompose the model parameters $\psi$ into the shared weights $\psi_{Sh.}$ and task-specific ones $\{\psi_{Pri.}, \psi_{Aux.1}, \psi_{Aux.2}\}$ for each branch.
To enable the model parameters to be customized according to the unique distribution of test samples, we propose to use meta-auxiliary learning to create the adapted parameters. 
Concretely, inspired by \cite{liu2019self,chi2021test}, our meta-auxiliary learning intends to learn the consistency of the parameters of our Aux. branches for the Pri. task, to ensure that the auxiliary tasks improve the performance of the Pri. task.
In the inner loop of the meta-training phase, several gradient updates of the auxiliary losses are used to update the parameter of the whole network parameter $\psi$, thereby performing effective adaptation on a specific sample.
Given a training pair $(\gtx^{(k)},\predx^{(k)})$, concerning the corrupted and repaired sequence, and $(p^{(k)},\td{p}^{(k)})$, w.r.t., the correct label of the disordered counterpart and the predicted one, it can be achieved:
\begin{small}
  \begin{equation}
    \td{\psi}^{(k)}\!\!  \leftarrow \psi \!-\!\alpha\nabla\! _{\psi}\left[\!\mathcal{L}_{Aux.1}(p^{(k)}\!,\td{p}^{(k)}\!)\! + \!\mathcal{L}_{Aux.2}(\gtx^{(k)\!},\predx^{(k)}\!)\!\right]\!\!
    \label{e11}
    \vspace{.5em}
  \end{equation}
\end{small}

\vspace{-1em}
\noindent where $\psi^{(k)} = \{\psi^{(k)}_{Sh.}, \psi_{Pri.}^{(k)}, \psi_{Aux.1}^{(k)}, \psi_{Aux.2}^{(k)}\}$ is the adapted parameter that is tailored by the specific observation.
$\alpha$ is the learning rate of the adaptation procedure.
We notice that both Aux. branches have the same gradient descent direction and are optimized concurrently to ensure a synergistic impact for the adaptation.

Our approach strives to maximize the performance of the Pri. forecasting branch by adjusting the model parameters through the self-supervised auxiliary tasks.
For this purpose, our meta-objective is formally denoted as:
\vspace{-.4em}
\begin{equation}
  \min _{\psi_{Sh.}, \psi_{Pri.}} \sum^{K}_{k=1}  \mathcal{L}_{Pri.}\left({\tb{Y}}_{1:\Delta T}^{(k)}, \td{\tb{Y}}_{1:\Delta T}^{(k)}; \td{\psi}^{(k)}_{Sh.}, \td{\psi}^{(k)}_{Pri.}\right).
  \vspace{-.6em}
  \label{e12}
\end{equation}
Here, $\mc{L}_{Pri.}$ is computed suing the pair $(\gtx^{(k)},\gty^{(k)})$, while the optimization is over $\psi = \{\psi_{Sh.},\psi_{Pri.}, \psi_{Aux.1}, \psi_{Aux.2}\}$ to achieve the updated parameter of the Pri. task.
Eq.\ref{e12} can be minimized using gradient descent algorithms:
\vspace{-.6em}
\begin{small}
  \begin{equation}
    \psi\leftarrow \psi -\beta\sum^{K}_{k=1} \nabla_{\psi} \mathcal{L}_{Pri.}\left({\tb{Y}}_{1:\Delta T}^{(k)}, \td{\tb{Y}}_{1:\Delta T}^{(k)}; \tilde{\psi}^{(k)}_{Sh.}, \tilde{\psi}^{(k)}_{Pri.}\right),
    \vspace{.3em}
  \end{equation}
\end{small}

\vspace{-1em}
\noindent where $\beta$ is the meta-learning rate.
The overall meta-auxiliary learning procedure is conducted in Algorithm.\ref{alg:a1}, in which the parameters of the Pri. task are updated in the outer loop, and the auxiliary parameters are updated in the inner loop. 
Regarding the testing phase, for a specific sequence, Eq.\ref{e11} is directly used to obtain the adapted parameters $\psi$, and then $\{\psi_{Sh.},\psi_{Pri.}\}$ is used to improve the generalization capability of the primary forecasting task.

\subsection{Implementation Details}
As shown in Fig.\ref{f2}, our model includes a Pri. and two Aux. branches.
The shared parts consist of $9$ residual blocks, created by combining the outputs of SS-RT and TS-RT, and having the channel $C_{in} = C_{out}=512$.
In addition, the task-specific portions of the Pri. and Aux.2 are an additional block to map the feature into the original dimension.
By contrast, the Aux.1 is a binary classifier, where its separate parts comprise a flatten layer, and 4 FC layers with channel numbers $256, 128, 64, 1$.
Aux.1 takes a scrambled-order counterpart of the observation as the input, while for Aux.2, we randomly remove 20\% of the joints from observations.
To reduce the complexity, in a specific layer, the GSU is shared in terms of spatio-temporal features.
Note that, the feature of the last layer of Pri., is directly connected to Aux.2, and passing through a flatten layer, is connected to Aux.1, so that the meta-auxiliary learning can update the whole parameters of the Pri. branch. 
We follow the current multi-task learning framework, and exploit the Adam optimizer to train our network, where the learning rate is initialized to $0.001$, with a $0.98$ decay every $2$ epoch.
The mini-batch size is $16$.
At the test-time adaptation, we fix the learning rate $\alpha = \beta = 2\times 10^{-5}$, and 6 gradient descents of Eq.\ref{e11} are performed.
Finally, the fine-tuned parameters are acquired, allowing for the adaptation of the internal properties of a specific sequence to achieve a better prediction, as shown in Fig.\ref{f1}.
Our code will be publicly available.

\begin{algorithm}[t]
  \caption{Meta-Auxiliary Training}
  \label{alg:a1}
  \tb{Require:} learning rates $\alpha$, $\beta$. pre-trained parameter $\psi = \{\psi_{Sh.},\psi_{Pri.}, \psi_{Aux.1}, \psi_{Aux.2}\} $  \\
  \tb{Ouput:} meta-auxiliary learned parameter\\ 
  \tb{1:} initialize the model with the pre-trained parameter $\psi$\\
  \tb{2:} \tb{while} \ti{not converge} \tb{do} \\
  \tb{3:} \quad sample a training batch from the $\{\gtx^{(k)},\gty^{(k)}\}_{k=1}^K$ ;\\
  \tb{4:} \quad \tb{for} \ti{each k} \tb{do} \\
  \tb{5:} \quad \quad evaluate the auxiliary losses $\mc{L}_{Aux.1},\mc{L}_{Aux.2}$; \\
  \tb{6:} \quad \quad update the adapted parameter: \\
  {}{}{\qquad}\small  ${\qquad}{\qquad}\!\!\! \td{\psi}^{(k)} \!\!=\! \psi \!-\!\alpha\nabla\! _{\psi}[\!\mc{L}_{Aux.1}(p^{(k)},\td{p}^{(k)}\!) \! +\! \mc{L}_{Aux.2}(\gtx^{(k)}\!,\!\predx^{(k)})]\!$\\
  \tb{7:} \quad \tb{end} \\
  \tb{8:} \quad evaluate the primary task and update:\\
  \small$ {\qquad}  \psi\leftarrow \psi -\beta\sum^{K}_{k=1} \nabla_{\psi} \mathcal{L}_{Pri.}(\gty^{(k)}, \predy^{(k)}; \tilde{\psi}^{(k)}_{Sh.}, \tilde{\psi}^{(k)}_{Pri.})$\\
  \tb{9:} \tb{end}
  \end{algorithm}
  \setlength{\textfloatsep}{.5em}

\section{Experiments}

\subsection{Preliminaries}

\tb{Dataset-1:} \tb{H3.6M} \cite{Ionescu2014Human36MLS} involves 15 action categories performed by 7 professional human subjects (\ti{S\_1}, \ti{S\_5}, \ti{S\_6}, \ti{S\_7}, \ti{S\_8}, \ti{S\_9}, \ti{S\_11}).
Each pose is represented as a 17-joint skeleton ($N=17$), and the sequences are down-sampled to achieve 25 fps \cite{mao2019learning,ma2022progy}.

\tb{Dataset-2:} We also select 8 action categories from \tb{CMU MoCap}.
The pre-processing solution is consistent with the H3.6M dataset.
For both H3.6M and CMU MoCap, the proposed model is implemented where the length of the observed sequence is equal to the prediction ($T=\Delta T = 25$).

\tb{Baselines.} To assess the effectiveness of the proposed approach, the following 5 state-of-the-art (SoTA) methods are selected as our baselines, including LTD \cite{mao2019learning}, DMGNN \cite{li2020dynamic}, MSR \cite{dang2021msr}, ST-Tr \cite{aksan2021spatio}, and PGBIG \cite{ma2022progy}.
LTD resorts to GCN to analyze the motion sequence in the frequency domain.
DMGNN suggests using GCNs to encode the human topology and RNNs for decoding.
MSR expands the multi-scale variant of the LTD approach.
ST-Tr exploits the spatial-temporal transformer for human motion prediction.
PGBIG is a recently introduced algorithm to generate a virtual initial guess to increase the prediction accuracy.

\tb{Metric.}
We test our model using the Mean Per Joint Position Error (MPJPE) in millimeters, in accordance with earlier work \cite{ma2022progy,dang2021msr}.

\begin{table}[t]
  \centering
  \scriptsize
  \renewcommand\arraystretch{.9}
  \setlength{\tabcolsep}{.9mm}{
    \begin{tabular}{|c|c|c|c|}
    \hline
    Datasets &testing &training &purpose \\
    \hline
    \multirow{3}{*}{H3.6M}  &\tb{(\ti{\romannumeral1})} \ti{S\_5}   &\ti{S\_1,S\_6$\sim$S\_9,S\_{11}}  &general predictive ability\\
    \cline{2-4}
    &\tb{(\ti{\romannumeral2})} \ti{S\_x}   &other subjects  &predictive ability on unseen subjects\\
    \cline{2-4}
    &\tb{(\ti{\romannumeral3})} \ti{C\_x}   &other categories  & \multirow{2}{*}{predictive ability on unseen categories} \\
    \cline{1-3}
    CMU MoCap  &\tb{(\ti{\romannumeral4})} \ti{C\_x} &other categories   &\\
    \hline  
  \end{tabular}
  }
  \vspace{-1.em}
\caption{ Experimental setups. 
As in the typical approaches, the \tb{setup-(\ti{\romannumeral1})} is to evaluate the general predictive ability, while the \tb{setup-(\ti{\romannumeral2})(\ti{\romannumeral3})(\ti{\romannumeral4})} are newly designed to investigate the adaptability to out-of-distribution data.
}
\label{t1}
\end{table}

\tb{Experimental Setups.}
We use 3 alternative setups to analyze our model, as stated in Table \ref{t1}.
\tb{(a)} testing on \ti{S\_5}, while training on (\ti{S\_1}, \ti{S\_6}, \ti{S\_7}, \ti{S\_8}, \ti{S\_9}, \ti{S\_11}), for the general prediction, as same as the prior methods \cite{mao2019learning,li2020dynamic,li2020dynamic,li2020dynamic};
To verify the performance for out-of-distribution data, the following new strategies are exploited:
\tb{(b)} testing on \ti{S\_x}, training on the actions on the other subjects, for the adaptability on unseen subjects;
\tb{(c)} testing on \ti{C\_x}, training on the actions on the other categories, for the adaptability on unseen action categories.
The prefix \ti{S} indicates the \ti{subject}, and \ti{C} denotes the \ti{category}.
For fairness, we also apply the training/testing division in Table \ref{t1}, but the hyperparameters remain unchanged, to re-train the baselines.

\begin{table}[t]
  \renewcommand\arraystretch{.99}
  \centering
  \scriptsize
  \setlength{\tabcolsep}{.17mm}{
      \begin{tabular}{|c|ccccc|ccccc|ccccc|}
        \hline
      \multicolumn{1}{|c|}{} &\multicolumn{5}{c|}{{walking}}&\multicolumn{5}{c|}{{eating}}&\multicolumn{5}{c|}{{smoking}}\\
      \multicolumn{1}{|c|}{ms} &80&160&320&400&1000&80&160&320&400&1000 &80&160&320&400&1000   \\ \hline
     LTD
     &12.3 &23.0 &39.8 &46.1 &59.8 
     &8.4  &16.9 &33.2 &40.7 &77.8 
     &7.9  &16.2 &31.9 &38.9 &72.6 
     \\
     DMGNN
     &17.3 &30.7 &54.6 &65.2 &95.8 
     &11.0 &21.4 &36.2 &43.9 &86.7 
     &9.0  &17.6 &32.1 &40.3 &72.2 
     \\
     ST-Tr
     &18.5 &27.4&60.1 &67.3 &103.2 
     &12.2 &24.5&40.7 &47.0 &84.2 
     &9.4  &17.9&38.4 &42.8 &79.6 
     \\
     MSR
     &12.2       &22.7&38.6 &45.2 &63.0 
     &\ul{8.4}   &17.1&33.0 &40.4 &77.1 
     &\ul{8.0}   &16.3&31.3 &38.2 &71.6 
     \\
     PGBIG
     &\tb{10.2} &\ul{19.8} &\ul{34.5}&\ul{40.3} &\ul{56.4 }
     &\tb{7.0 } &\tb{15.1} &\ul{30.6}&\ul{38.1} &\ul{76.0 }
     &\tb{6.6 } &\ul{14.1} &\ul{28.2}&\ul{34.7} &\ul{69.5 }
     \\
     \hline
     Ours 
     &\ul{10.8}   &\tb{18.9} &\tb{33.2}&\tb{38.1} &\tb{52.3}
     &{8.8}       &\ul{15.4} &\tb{28.5}&\tb{36.7 }  &\tb{71.6} 
     &\tb{6.6}    &\tb{13.5} &\tb{26.7}&\tb{32.0 }  &\tb{67.5} 
    \\
  \hline
      \end{tabular}
  }
  \vspace{-1em}
\caption{MPJPE comparisons on 3 activities from the H3.6M dataset, where the experimental design follows the conventions of predictive algorithms(\ti{S\_5} is used for testing while the other is used for training).
The best result is displayed in bold, while the second is underlined.
We observe our model achieves the overall better results.
It reveals that the dynamic characteristics of \ti{S\_5} are slightly different from those of other subjects, and our approach is able to adapt to them.
}
\vspace{-.1em}
\label{t2}
\end{table}

\begin{table}[t]
  \centering
  \small
  \renewcommand\arraystretch{.95}
  \setlength{\tabcolsep}{1.4mm}{
    \begin{tabular}{|c|cccccc|}
    \hline
    Unseen Subjects &\ti{S\_1}& \ti{S\_6}& \ti{S\_7}& \ti{S\_8}&\ti{S\_9}&\ti{S\_11} \\\hline
    LTD      &115.4  &132.8   &133.7 &120.1 &123.8 &124.3\\
    DMGNN    &122.5  &139.3 &131.0 &125.2 &134.7 &120.2\\
    ST-Tr    &133.6  &147.5  &134.2 &128.0 &140.2 &124.5\\
    MSR      &115.7  &131.0  &\ul{123.1} &\ul{116.5} &118.8 &116.2\\
    PGBIG    &\ul{113.2}  &\ul{127.3}  &124.4 &118.3 &\ul{114.6} &\ul{112.0}\\
    \hline
    Ours  &\tb{107.0} &\tb{123.2} &\tb{118.7} &\tb{113.5} &\tb{109.7} &\tb{110.2} \\
    \hline
    \end{tabular}
  }
  \vspace{-1em}
\caption{Average MPJPE of a total of 15 activities at the end predicted pose (1000ms), evaluated on each unseen subject.
}
\label{t3}
\end{table}

\subsection{Comparison with State-of-the-arts on H3.6M}
\tb{General predictive ability.}
The existing predictors are normally tested on the actions of $\ti{S\_5}$ and trained on the other subjects.
However, our key observation is that, the motion patterns of different individuals tend to be distinct; therefore, this distribution-shift deteriorates the performance of deep pre-trained models.
As a comparison, at test-time, our model is able to be further optimized by meta-auxiliary learning, to achieve a better result.
Consistent with the previous work \cite{mao2019learning,cui2020learning}, we first use the \tb{setup-(\ti{\romannumeral1})} (in Table \ref{t1}) to evaluate the general predictive ability of our model.
Table \ref{t2} reports the comparison of 3 representative activities.
We observe that, our result tends to be better in almost all scenarios, which reveals that the dynamic characteristics of \ti{S\_5} are potentially distinguishing from other subjects, and our model can adapt to them.

\tb{Predictive ability on unseen subjects.}
Intuitively, due to unique height and body proportion, even for the same category, the motion properties (\eg, styles or rhythms) of different subjects are potentially inconsistent.
To further investigate the adaptation ability of different subjects, the experimental \tb{setup-(\ti{\romannumeral2})} is used.
Concretely, we fine-tune the base model under the test actions of a specific subject-x (\ti{S\_x}), where the base model is learned from the others.
Table \ref{t3} provides the average MPJPE of the end predicted pose (1000ms) of different unseen subjects.
From the results, we observe that our model produces better predictions against the baseline models.
It implies that the dynamic characteristics of different humans indeed involve distinct motion attributes.
Moreover, our approach exploits the external large dataset, and meanwhile, can be tailored based on the internal information of test sequences via meta-auxiliary learning, to consistently yield a superior result for unseen subjects.

\begin{figure*}[htbp]
  \centering
  \includegraphics[width=6.85in,height=2.9in]{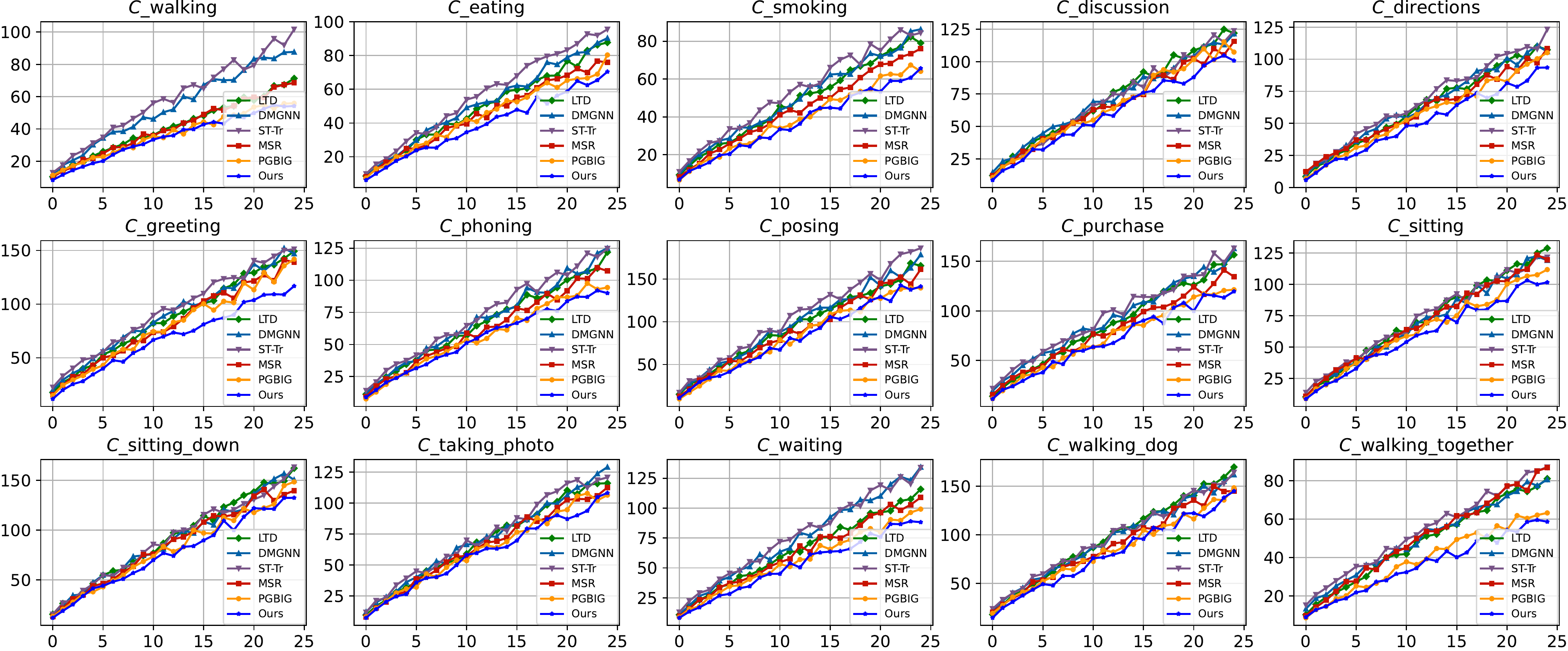}
  \vspace{-1.em}
  \caption{Comparison of each unseen action category \ti{C\_x} from the H3.6M dataset.
  We observe that, at test-time, our approach is able to be fine-tuned for a specific category \ti{C\_x} to adapt to its internal properties, thus achieving the higher prediction accuracy.
  }
  \label{f4}
\end{figure*}

\begin{figure*}[!t]
  \centering
  \includegraphics[width=6.92in,height=1.5in]{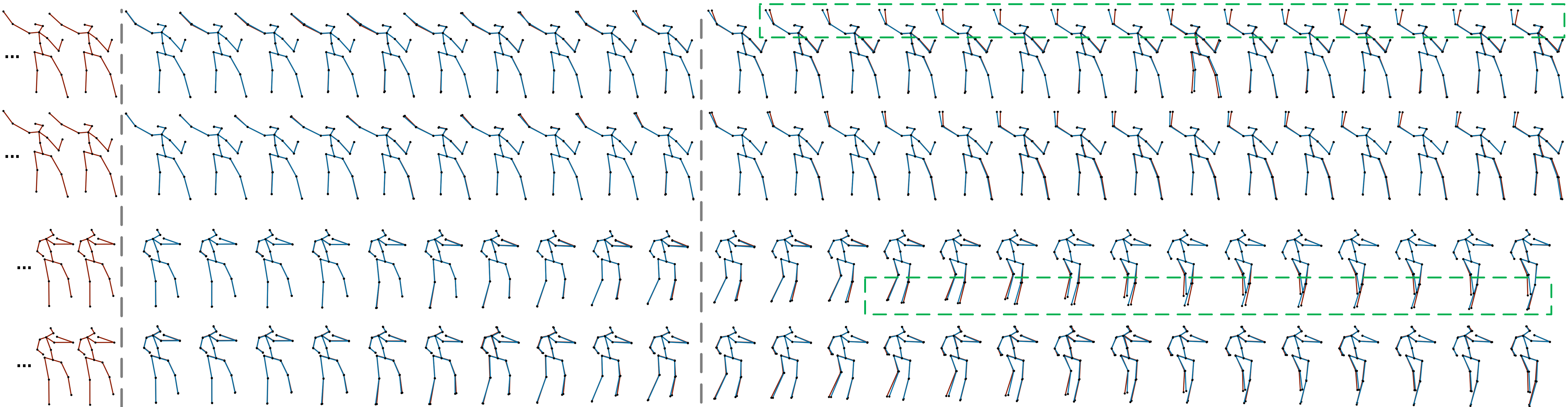}
  \vspace{-.5em}
  \caption{\small Qualitative comparison on the greeting activity of unseen subjects \ti{S\_11} (top) and unseen category \ti{C\_phoning} {(bottom)}.
  In each sub-figure, the first row is the SoTA PGBIG \cite{ma2022progy}, followed by our result, where the blue pose refers to the prediction, and the underlying red is the GT.
  The green rectangles indicate the contrasting parts.
  We observe that, our predicted poses are closer to the GT, as it is tailored according to the specific sequence.
  }
  \vspace{-.5em}
  \label{f5}
\end{figure*}

\tb{Predictive ability on unseen categories.}
Due to the diversity and uncertainty, human action involves unenumerable categories.
Typically, the training dataset falls short of covering all action types.
In practical applications, existing deep end-to-end algorithms face a major challenge, that is, once encountering the unseen category at test-time, their performance tends to decline sharply.
However, our model is able to further optimize the base model learned from large datasets, to adapt to the unique attributes of a new action category.
To verify it, we exploit the experimental \tb{setup-(\ti{\romannumeral3})}.
Specifically, our approach and the baselines are evaluated under each specific category \ti{C\_x} respectively, while the training is conducted on the remaining ones.
From Fig.\ref{f4}, we observe that our model brings superior results in all scenarios for such out-of-distribution data of unprecedented categories.
It evidences that our model is indeed capable of adapting to the characteristics of unseen action categories.

Also, Fig.\ref{f4} illustrates two qualitative comparisons between the proposed model and the SoTA PGBIG \cite{ma2022progy}, for the \ti{greeting} activity of the unseen subject-11 (\ti{S\_{11}}) and unseen action category (\ti{C\_{phoning}}).

\begin{figure}[htbp]
  \centering
  \includegraphics[width=3.2in,height=1.3in]{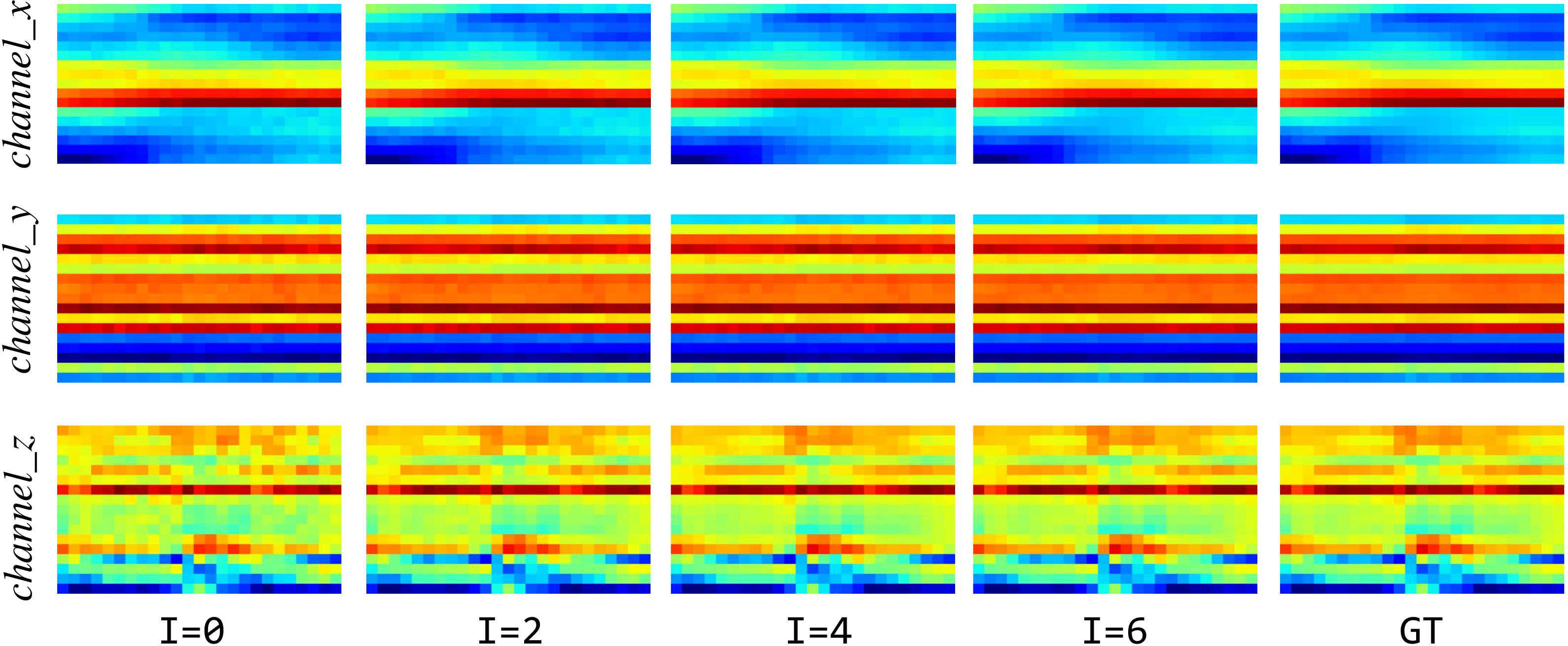}
  \vspace{-.5em}
  \caption{\small Results of the unfolding TTA process with the different number of gradient descents $I=\{0,2,4,6\}$ on \ti{C\_smoking}.
  With the iteration, the coarse results tend to be close to the GT.
  }
  \label{f6}
\end{figure}

\tb{Results on CMU MoCap.}
We also evaluate the predictive ability on unseen categories from the CMU MoCap using the \tb{setup-(\ti{\romannumeral4})}.
From Fig.\ref{f5}, the results show that our model substantially outperforms the baselines.

\begin{table}[H]
  \centering
  \scriptsize
  \renewcommand\arraystretch{.83}
  \setlength{\tabcolsep}{.2mm}{
      \begin{tabular}{|c|cccccccc|}
        \hline
        \rotatebox{45}{{\tabincell{c}{Unseen \vspace{-0.3em} \\ Categories}}}&\rotatebox{45}{\ti{C\_{basket}}}&\rotatebox{45}{\ti{C\_{bas-sig}}}
        &\rotatebox{45}{\ti{C\_{dir-trc}}}
        &\rotatebox{45}{\ti{C\_{jumping}}} &\rotatebox{45}{\ti{C\_{running}}}&\rotatebox{45}{\ti{C\_{soccer}}}&\rotatebox{45}{\ti{C\_{walking}}}&\rotatebox{45}{\ti{C\_{washwin}}} \\ \hline
      LTD       &109.0      &75.3      &121.2      &142.4      &65.7        &115.3     &49.0         &83.1 \\
      DMGNN     &145.6      &72.7      &130.3      &163.1      &{73.4}   &121.9     &{52.1}    &{89.8} \\
      ST-Tr     &150.2      &77.1      &131.0      &153.2      &{76.8}   &130.4     &{60.3}    &{95.6} \\
      MSR       &{101.1} &{66.6} &{117.4} &{138.6} &\ul{56.3}   &{110.4}&45.2        &74.9 \\ 
      PGGIG     &\ul{94.9}  &\ul{61.9} &\ul{113.0} &\ul{134.8} &57.1        &105.1     &\ul{41.4}   &74.7  \\ \hline
      {Ours }   &\tb{87.4}  &\tb{53.1} &\tb{97.3}  &\tb{120.4} &\tb{46.2}   &\tb{93.2} &\tb{35.8}   &\tb{62.5} \\
      \hline
      \end{tabular}
  }
  \vspace{-1.em}
\caption{ \small  {Average MPJPE of the end predicted pose (1000ms) of each unseen category \ti{C\_x} from the CMU MoCap dataset.}
}
\vspace{-1.em}
\label{t4}
\end{table}

\tb{Progressive results. }
At test-time, with several gradient updates, our model allows us to learn to adapt to the internal properties of the test sequence.
To better explain it, we show the progressive result by unfolding the TTA procedure after each gradient descent.
The inference is run on \ti{C\_smoking}, and the training is run on the other categories.
Fig.\ref{f6} presents the 3 channels (\ie, x, y, z axes) of these intermediate results by the heat map, with more red denoting larger, and more blue, smaller values.
We see that as the iteration goes on, the result gradually tends to be closer to the GT.

\subsection{Ablation Studies}
Here, the following ablation experiments are conducted.
We adapt our approach to each action category \ti{C\_x} and take the average as the result, as in the \tb{setup-(\ti{\romannumeral3})}.

\tb{\ti{w/} GSUs \ti{v.s.} \ti{w/o} GSUs.}
Intuitively, the GSU facilitates the transfer of useful information.
It is confirmed in Table \ref{t5}.

\begin{table}[t]
  \centering
  \small
  \vspace{-.2em}
  \renewcommand\arraystretch{.85}
  \setlength{\tabcolsep}{2.9mm}{
      \begin{tabular}{|c|cccc|}
        \hline
      \multicolumn{1}{|c|}{Millisecond (ms)}  &80 &160 &400 &1000  \\ \hline
       \ti{w/o} GSUs     &11.2      &23.1      &60.3      &109.3  \\
       \ti{w/} GSUs      &\tb{10.1} &\tb{21.6} &\tb{55.2} &\tb{104.1} \\
      \hline
      \end{tabular}
  }
  \vspace{-1em}
\caption{{Effects of the proposed GSU.}
}
\vspace{-.2em}
\label{t5}
\end{table}

\tb{Impact of Aux. branches.}
Both the Aux.1 and Aux.2 branches behave as the complement to the Pri. task.
To verify the effectiveness of the Aux. branches, we analyze the effects to the Pri. of the Aux.1 and Aux.2 on retaining one of them.
As shown in Table \ref{t6}(left), when Aux.1 and Aux.2 are introduced concurrently, a better result is achieved.

\tb{Number of gradient descents.}
Here, we provide the impact of the maximum number of gradient updates $I=\{0,5,6,7\}$ at test-time adaptation.
From Table \ref{t6}(right), we observe that, overall, the larger $I$ obtains smaller errors.
When $I=5$, the best result is yielded, and larger value brings no benefits.

\begin{flushleft}
\begin{table}
  \flushleft
  \begin{tabular}{l}
    \!\! \!\! \!\! \!\! \!\! \!\! \!\! \!\! \!\! \!\! \!\! \!\!
    \small
    \renewcommand\arraystretch{1.4}
    \setlength{\tabcolsep}{.08mm}{
    \begin{tabular}{|cc|cccc|}
      \hline
    \ti{Aux.1} &\ti{Aux.2}  &80 &160  &400 &1000   \\ \hline
    \checkmark &$\times$    &12.8 &25.2  &62.7   &111.0   \\
        $\times$   &\checkmark  &11.4     &22.6 &58.5   &108.4       \\
        \checkmark &\checkmark  &\tb{10.1} &\tb{21.6} &\tb{55.2} &\tb{104.1} \\
    \hline
    \end{tabular}
    }
    \renewcommand\arraystretch{1.2}
    \setlength{\tabcolsep}{1mm}{    
    \begin{tabular}{|c|cccc|}
          \hline
        \multicolumn{1}{|c|}{}  &80 &160 &400 &1000  \\ \hline
        $\!\!I\!=\!0\!$      &10.7 &23.2 &58.1 &109.5  \\
        $\!\!I\!=\!5\!$      &10.3 &22.1 &56.0 &{106.2} \\
        $\!\!\bs{I\!=\!6}\!$ &\tb{10.1} &\tb{21.6} &\tb{55.2} &\tb{104.1}\\
        $\!\!I\!=\!7\!$      &{10.7} &22.0 &\tb{54.9}     &106.7\\
        \hline
        \end{tabular}
    }
    \end{tabular}
  \vspace{-1.em}
  \caption{\small  Effects of the Aux. branches (left), and different number of gradient descents (right). $I=0$ indicates the base model.}
  \label{t6}
\end{table}
\end{flushleft}

\section{Conclusion}
In this work, we have introduced a test-time adaptation model for human motion forecasting.
It uses meta-auxiliary learning to ensure that the update of auxiliary tasks can bring superior adaptability and better performance to the main task on specific samples. 
At test-time, it resorts to meta-auxiliary learning to ensure that the updates of both auxiliary tasks can bring better adaptation and higher performance to the primary task on specific sequences.
Extensive experiments show that our model consistently outperforms the SoTA approaches on unseen subjects and categories.
It has revealed that our model is able to adapt to the dynamic characteristics of out-of-distribution data in the real world.

\section*{Acknowledgements}
This work was supported in part by the National Natural Science Foundation of China (62176125), in part by the Jiangsu Funding Program for Excellent Postdoctoral Talent (2022ZB269), in part by the Natural Science Foundation of Jiangsu Province (BK20220939), and in part by the China Postdoctoral Science Foundation (2022M721629).
\bibliography{aaai23}

\end{document}